\documentclass[runningheads]{llncs}
\usepackage[T1]{fontenc}
\usepackage{graphicx}
\usepackage[acronym]{glossaries}
\makeglossaries

\newacronym{ml}{ML}{Machine Learning}
\newacronym{llm}{LLM}{Large Language Model}
\newacronym{sota}{SotA}{state-of-the-art}
\newacronym{iaa}{IAA}{inter-annotator agreement}
\newacronym{docvqa}{DocVQA}{Document Visual Question Answering}
\newacronym{lv}{LV}{Language-Vision}
\newacronym{dia}{DIA}{Document Image Analysis}
\newacronym{vqa}{VQA}{Visual Question Answering}
\newacronym{idvqa}{iDocVQA}{Instruction Document Visual Question Answering}
\newacronym{pali}{PaLI}{Pathways Language and Image}
\newacronym{clip}{CLIP}{Contrastive Language–Image Pre-training}
\newacronym{llava}{LLaVA}{Large Language and Vision Assistant}
\newacronym{llama}{LLaMA}{Large Language Model Meta AI}
\newacronym{lmm}{LMM}{Large Multimodal Model}
\newacronym{mlp}{MLP}{multilayer perceptron}
\newacronym{beit3}{BEiT3}{BERT Pre-Training of Image Transformers-3}
\newacronym{nlp}{NLP}{natural language processing}
\newacronym{cv}{CV}{computer vision}
\newacronym{donut}{Donut}{Document Understanding Transformer}
\newacronym{cnn}{CNN}{Convolutional Neural Network}
\newacronym{ocr}{OCR}{optical character recognition}
\newacronym{ie}{IE}{information extraction}
\newacronym{ir}{IR}{information retrieval}
\newacronym{cord}{CORD}{consolidated receipt dataset}
\newacronym{llad}{LLaDoc}{Large Language Document}
\newacronym{squad}{SQuAD}{Stanford Question Answering Dataset}
\newacronym{spdvqa}{SP-DocVQA}{Single Page Document Visual Question Answering}
\newacronym{tvqa}{TextVQA}{Text Visual Question Answering}
\newacronym{pope}{POPE}{Polling-based Object Probing Evaluation}
\newacronym{CoT}{COT}{Chain-of-Thought}
\newacronym{LMs}{LM}{Language models}
\newacronym{wandb}{WandB}{weights and biases}

\begin{document}
\title{Instruction Makes a Difference}

%
%

\author{Tosin Adewumi\inst{*} \and
Nudrat Habib\inst{} \and
Lama Alkhaled\inst{} \and
Elisa Barney\inst{}}
\authorrunning{Adewumi et al.}
%
\institute{Machine Learning Group, EISLAB, Luleå University of Technology, Sweden. \\
\email{firstname.lastname@ltu.se}\\
*corresponding author
}
\maketitle              

\begin{abstract}
We introduce the \acrfull{idvqa} dataset and the \acrfull{llad} model, for training \acrfull{lv} models for document analysis and predictions on document images, respectively.
Usually, deep neural networks for the \acrshort{docvqa} task are trained on datasets lacking instructions.
We show that using instruction-following datasets improves performance.
We compare performance across document-related datasets using the recent \acrfull{sota} \acrfull{llava}1.5 as the base model.
We also evaluate the performance of the derived models for object hallucination using the \acrfull{pope} dataset.
The results show that instruction-tuning performance ranges from 11x to 32x of zero-shot performance and from 0.1\% to 4.2\% over non-instruction (traditional task) finetuning.
Despite the gains, these still fall short of human performance (94.36\%), implying there's much room for improvement.

\end{abstract}
\section{Introduction}
\label{intro}
The task of \acrfull{docvqa} involves natural language answers to questions based on document images \cite{mathew2021docvqa,10.1007/978-3-030-86337-1_42}.
The example of the modern trend of using smart phones to capture and save documents or mixed image-text content makes \acrfull{dia} and its sub-tasks even more relevant today.
Such documents are more challenging than digitally-created documents in \acrshort{dia} \cite{xu2020layoutlm}.
According to \cite{aiim2023}, about 25\% of professionals in the U.S. do not think paper invoices within organizations will be eradicated by 2025 while another 25\% are unsure.

The convergence of \acrfull{nlp} and \acrfull{cv} has been accelerating in recent times \cite{chen2022pali,dai2023instructblip,liu2023llava,wang2023image}.
This has been facilitated by the use of the Transformer architecture \cite{vaswani2017attention} in the \acrshort{nlp} domain, which has gained attention in the \acrshort{cv} domain \cite{dosovitskiy2020image,yuan2021tokens}.
As identified by \cite{mishra-etal-2022-cross}, cross-task generalization in a \acrfull{llm} benefits from datasets with instructions \cite{touvron2023llama}.
This cross-task generalization involves learning a model that at inference (without previous task-specific training) produces a specific output based on specific input and task instruction.
An instruction dataset is one that contains direction on how a task should be done.

\acrfull{lv} instruction-(fine)tuning is under-explored \cite{dai2023instructblip}.
Typically, the training of a \acrfull{lmm} (i.e. with more than one modality) has two steps.
The first is the alignment pretraining that aligns the visual features from the images fed to the vision encoder with the language model’s word embedding space \cite{liu2023improvedllava}.
The second step is the instruction-tuning that allows the model to follow a user's directives.
Integrating \acrshort{llm}s with vision components can bring many benefits.
Such \acrshort{lmm}s can do more than just QA tasks, such as summarizing a document, having multiturn conversations about a document, or writing poems based on a document or an image \cite{fu2023mme,info13060298}.
Hence, incorporating vision or additional modalities to \acrshort{llm}s will help them to solve novel tasks \cite{alayrac2022flamingo}.
Instruction-tuning improves the zero-shot capabilities of \acrshort{llm}s \cite{liu2023llava} but there appears to be a gap on their impact in \acrshort{docvqa}, as many of the existing datasets are designed as simple question-answer pairs.
The research question we, therefore, address is simple: `\textit{How effective is an instruction dataset, in the \acrshort{lmm} context, for improving the performance of \acrshort{docvqa}}?'.

We show that improvements are possible in a series of experiments involving 3 datasets: \acrfull{docvqa} \cite{mathew2021docvqa}, \acrfull{tvqa} \cite{8953586}, and \acrfull{idvqa}.
The instruction we use contain key ingredients: \textbf{task name}, a \textbf{persona} for the \acrshort{llm}, and the \textbf{type of output} desired.
We adopt the \acrfull{sota} \acrfull{llava}1.5-7B model and evaluate it in different scenarios, including (1) zero-shot (baseline), (2) traditional task (non-instruction) finetuning, (3) instruction-tuning and (4) 50-50 instruction-task tuning.
We further evaluate the derived models for object hallucination on the \acrfull{pope} benchmark dataset \cite{li-etal-2023-evaluating}. 
The following are our contributions in this work.

\begin{enumerate}
    \item We create and publicly release a new multimodal English instruction dataset\footnote{github.com/LTU-Machine-Learning/iDocVQA} dataset
    \item We publicly release our \textbf{\acrfull{llad}\footnote{huggingface.co/tosin/LLaDoc}} model\textsuperscript{1} - an \acrshort{lmm} which is also multilingual, based on \acrfull{llama}-2.
    \item We show through experiments and analysis that well-written instructions improve performance on the \acrshort{docvqa} task.
\end{enumerate}

The rest of this paper is organized as follows.
We present a literature review of related work in Section \ref{lit}, including \acrshort{llm}s and harmful content, which they are prone to produce sometimes.
Section \ref{method} describes details of the methodology, including the dataset creation and the architecture of \acrshort{llad}.
In Section 4,
we present the findings, the analysis of the results, and some qualitative examples.
We conclude with our final thoughts and possible future work in Section 5. 

\section{Literature Review}
\label{lit}

A recognition-free approach to \acrshort{docvqa} by \cite{mathew2021asking} was evaluated on two datasets, including HW-SQuAD, a synthetic image version of version 1 of the benchmark \acrfull{squad} \cite{rajpurkar-etal-2016-squad} to pass off as handwritten documents.
Their approach focused on visual evidence as a form of explanation for an answer.
This approach is supported by the STE \acrshort{vqa} dataset, in addition to the provision for textual answer \cite{wang2020general,mathew2021asking}.
None of the foregoing datasets are designed as instruction datasets.
Creating visual instruction-tuning data usually involves adding instructions to existing image captioning and \acrshort{vqa} data \cite{zhao2023svit,liu2023llava}, as there is typically high cost associated with creating new datasets of high volume, especially when they involve multiple or low-resource languages \cite{10191208}.

\cite{10.1007/978-3-030-86337-1_42} makes a distinction between Single Document and Document Collection \acrshort{vqa} tasks and introduced the Infographics \acrshort{vqa} task based on the dataset with 5,485 infographics images.
Many efforts at \acrshort{vqa}, such as LayoutLM \cite{xu2020layoutlm} and \acrfull{donut} \cite{kim2022donut}, consist of
deep image features, a question embedding module, and fusion of the image and text modalities \cite{anderson2018bottom,teney2018tips,10.1007/978-3-030-86337-1_42,hu2020iterative}.
Earlier efforts leaned towards \acrfull{cnn}-based architectures to detect structures in documents \cite{hao2016table,he2017mask,ren2015faster}.
The Transformer architecture now plays an important roll in similar efforts.
In \cite{xu2020layoutlm}, they focus on layout and style of documents as additional features to improve \acrshort{dia}.
\cite{hu2020iterative} extracts feature representations from the  question words, visual objects, and \acrshort{ocr} tokens before projecting them into the same semantic space.
\acrshort{donut}, on the other hand, maps raw image inputs to the desired outputs without \acrfull{ocr} \cite{kim2022donut}.
Indeed, methods that use multimodal pretrained architectures have been shown to outperform those based solely on language representations \cite{10.1007/978-3-030-86337-1_42}.

Recent efforts, such as the \acrfull{pali} model by \cite{chen2022pali}, \acrfull{beit3} \cite{wang2023image}, and \acrfull{llava} \cite{liu2023llava}, combine natural language models with computer vision encoders for wider understanding capabilities.
In \cite{chen2022pali}, scaling up the language and vision unimodal components saves compute and improves performance to achieve \acrshort{sota} on various tasks.
\acrshort{pali} combines a visual Transformer, a multimodal encoder and a text decoder.
\acrshort{beit3} performs, in a unified approach, masked language modeling on images, English texts, and image-text pairs.
Its backbone is the Multiway Transformer, which has layers of switching modality experts, consisting of feedforward networks for language, vision and vision-language.

In \cite{liu2023llava}, they use a projection matrix and combine pre-trained \acrfull{clip} ViT-L/14 visual encoder and Vicuna \acrshort{llm}, similarly to \cite{zhu2023minigpt}, which is based on \acrfull{llama} \cite{touvron2023llama}.
\acrshort{clip} builds on natural language supervision, zero-shot transfer, and multimodal learning \cite{radford2021learning}.
Improving on \cite{liu2023llava}, \cite{liu2023improvedllava} uses a two-layer \acrfull{mlp} instead of the linear projection in the former.
Among the limitations in \cite{liu2023improvedllava} is the inability of the model to process multiple images because of context size and the lack of such instruction dataset in its training.

Instructions, such as \acrfull{CoT} prompting \cite{wei2022chain}, assists  \acrshort{llm}s in performing complex tasks when prompted with a few exemplars, providing stepwise solutions on such tasks thus enhancing their reasoning.
Experimental results using multiple \acrshort{llm}s have shown improved performance on various tasks with PaLM being the best-performing model, achieving new \acrshort{sota} results on GSM8K, SVAMP, MAWPS, and strategyQA \cite{wei2022chain}.
Designing effective \acrshort{CoT} prompts requires human experts with an understanding of both the task and the prompting technique, which limits its scalability and generalizability.
Another study showing the zero-shot capability of \acrshort{llm}s by simply adding “Let’s think step by step” was performed by \cite{kojima2022large}.  
Zero-shot \acrshort{CoT} achieved massive score gains compared to a zero-shot baseline on diverse benchmark reasoning tasks.
To reduce the cost and dependency on humans for step-by-step thought generation and better generalization, a reprompting algorithm was designed \cite{xu2023reprompting}.
It is an iterative sampling algorithm that searches for the \acrshort{CoT} recipes for a given task without human intervention.
It achieves consistently better performance than the zero-shot, few-shot, and human-written \acrshort{CoT} baselines \cite{xu2023reprompting}. 

\subsection*{LLM datasets to address harmful content}
The study by \cite{wang2023not} suggests that \acrshort{llama}-2 is the safest model when prompting \acrshort{llm}s for risky or harmful outputs, followed by ChatGPT, Claude, GPT-4, and Vicuna.
Harmful content refers to content that could be offensive, misleading, or negatively impactful, such as hallucination \cite{li-etal-2023-evaluating,adewumi2023procot}.
The presence of harmful content in the outputs of \acrshort{llm}s is a significant concern because it can perpetuate and amplify social issues, including discrimination, polarization, and misinformation.
The real toxicity prompts dataset was developed as an early work to facilitate research into the safety alignment of \acrshort{llm}s \cite{gehman2020realtoxicityprompts}.
Bias Benchmark for Question Answering (BBQ) is another collection of questions to expose and examine social biases within protected groups across 9 key aspects, particularly in contexts where U.S. English is spoken \cite{parrish2022bbq}.
Research using LMSYS-Chat-1M \cite{zheng2023lmsys}, a comprehensive dataset comprising one million conversations involving 25 \acrshort{llm}s, indicates that numerous conversations with potential harm were not identified by OpenAI's moderation API. On the other hand, \acrshort{llama}-2-7B-Chat tends to decline most moderation-related prompts, possibly due to an overly cautious stance towards harmful content.


\section{Methodology}
\label{method}

\begin{figure}[h]
\centering
\includegraphics[width=0.9\textwidth]{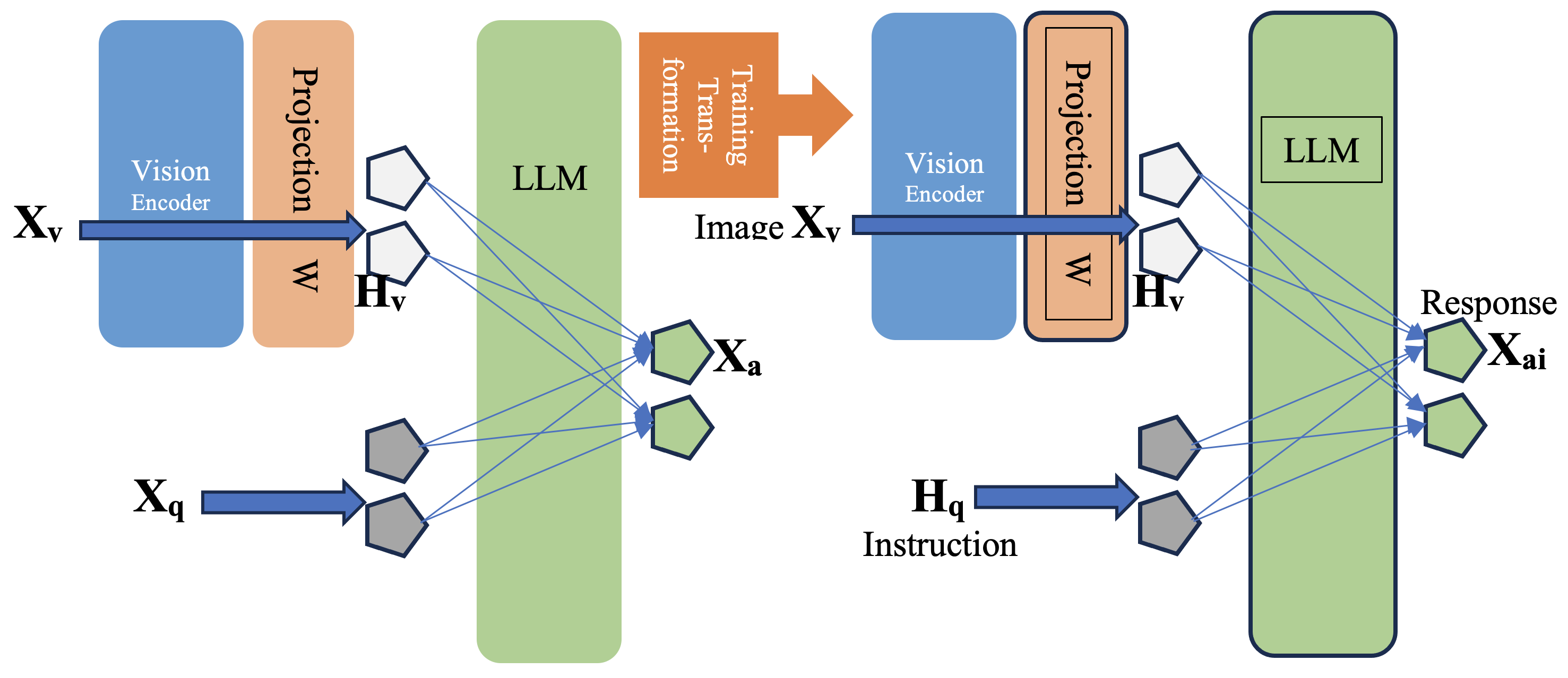}
\caption{\acrshort{llad} architecture/training schema} \label{archi}
\end{figure}

Details about the \acrshort{idvqa} dataset creation are provided in the next subsection.
All the experiments were conducted on a single node of 8 NVIDIA A100-SXM 40GB GPUs, running Ubuntu 22.04 and CUDA 12.3 with FlashAttention-2 \cite{dao2023flashattention2}.
Each experiment was run twice (because of computation cost) and the average score recorded, including standard deviation.
We only evaluated the validation sets in all cases\footnote{due to resource constraints}, which is usually indicative of the test set performance.
Training time ranged from about 3.1 to 6.9 hours, depending on the data size, while evaluation time ranged from about 15 to 90 minutes, depending on the model and data size.

We perform full parameter tuning of the models using \acrshort{llava}-1.5-7B on the \acrshort{docvqa} \cite{mathew2021docvqa}, \acrshort{tvqa} \cite{8953586}, and \acrshort{idvqa} datasets.
This is done in 3 ways: non-instruction finetuning, full instruction-tuning, and ablation study of using 50-50\% of instruction and no instruction in the training and evaluation sets.
In addition, we perform direct inference on the baseline model (\acrshort{llava}1.5).
We call the best performing model checkpoint trained on the \acrshort{idvqa} dataset \acrshort{llad}, which emerges as a result of the weight changes to the \acrshort{mlp} and the \acrshort{llama} \acrshort{llm}.
Figure \ref{archi} presents the schema of the architecture/training. \acrshort{llava} is based on \acrshort{llama} and OpenAI CLIP-ViT.
We chose \acrshort{llava}1.5 because it is \acrshort{sota} and many other \acrshort{lmm}s are based on it \cite{chen2023sharegpt4v,zhou2023infmllm}.

Our protocol for all the training involves the use of the CLIP-ViT large as vision encoder because it is currently the best performing model, 12 training \textit{epochs}, \textit{batch size} of 32, \textit{gradient accumulation step} of 1, initial \textit{learning rate} of 2e-5, no weight decay, \textit{warm up ratio} of 0.03, \textit{cosine} learning rate scheduler, and maximum \textit{context length} of 2,048.
During evaluation, we use the default \textit{temperature} of 0.2.
In line with previous work, we report accuracy scores \cite{hu2023bliva,mathew2021docvqa,10.1007/978-3-030-86337-1_42} and enforce that correct answers must be an exact match of the ground truth.
We also evaluate the derived models on the following benchmark dataset for harmful hallucination generation:

\begin{itemize}
    \item \acrfull{pope} \cite{li-etal-2023-evaluating}.
    It is a polling-based query approach that systematically evaluates object hallucination of \acrshort{lv} models.
    We evaluate with a temperature of 0 under the three settings: random, popular and adversarial.
    It evaluates hallucination through binary classification by prompting models with Yes-or-No short questions.
    We report \textit{F1} and \textit{Yes} ratio and use the default \textit{temperature} setting of 0.
    \acrshort{pope} is limited in that it does not reflect overall performance of \acrshort{lmm}s and there aren't exact correlation always between the  \textit{F1} and \textit{Yes} ratio.

\end{itemize}



\subsection{\acrshort{idvqa} Dataset Creation}

We merged and transformed the \acrshort{docvqa} and \acrshort{tvqa} datasets into the \acrshort{idvqa} dataset for instruction-finetuning, resulting in a somewhat more challenging dataset because of the increased diversity.
The first 2 are similar in that \acrshort{docvqa} is a collection of single page printed, typewritten, handwritten and born-digital text with 50,000 questions while \acrshort{tvqa} is a collection of diverse images with text (such as billboards and traffic signs) consisting of 45,336 questions.
Table \ref{stati} gives statistics of the datasets and Table \ref{tabex} shows examples in the \acrshort{idvqa} training data in a tabular format.
We transformed the datasets into the JSON \acrshort{llava} format \cite{liu2023llava} in order to finetune with the model.
Each question and answer pair is formatted into the conversations field of the dataset.
We use a general system instruction for the entire dataset: `\textbf{\#\#\#Instruction: Following is a \acrfull{vqa} task. As a helpful system, give a suitable response as output, using the input for more context if it is provided:}'.
We believe this template is more effective because it provides a persona, identifies the task, and the desired type of output.
Part of the applicable rule of thumb we follow includes clear delimitation using `\#\#\#' and being as descriptive as possible in the instruction.\footnote{platform.openai.com/docs/guides/prompt-engineering}

\begin{table}[h!]
\centering
\caption{Statistics of the training \& validation sets, and total images.}\label{stati}
\begin{tabular}{p{0.27\linewidth} | p{0.2\linewidth} | p{0.2\linewidth} | p{0.2\linewidth}}
\hline
 &  \acrshort{idvqa} & \acrshort{docvqa} & \acrshort{tvqa}\\
\hline
Training set & 74,065  & 39,463 & 34,602 \\
Validation set &  10,349 & \hspace{0.09cm} 5,349 & \hspace{0.09cm} 5,000 \\
Total samples & 84,414 & 44,812 & 39,602 \\
\hline
Total images & 37,889  & 12,767 & 28,408\\
\hline
\end{tabular}
\end{table}

\begin{table}[h!]
\centering
\caption{Extracts of samples in the \acrshort{idvqa} training data.}\label{tabex}
\begin{tabular}{p{0.06\linewidth} | p{0.65\linewidth} | p{0.1\linewidth} | p{0.17\linewidth}}
\hline
ID & Question &  Answer & Image file\\
\hline
337-279 & \#\#\# Instruction: Following is a Visual Question Answering (VQA) task. As a helpful system, give a suitable response as output, using the input for more context if it is provided:what is the date mentioned in this letter?  & 1/8/93 & xnbl0037\_1.png\\
338-279 & \#\#\# Instruction: Following is a Visual Question Answering (VQA) task. As a helpful system, give a suitable response as output, using the input for more context if it is provided:what is the contact person name mentioned in letter? & P. Carter & xnbl0037\_1.png\\
\hline
\end{tabular}
\end{table}








\section{Results and Discussion}
\label{result}

Four methods were experimented with on the 3 datasets.
The results are presented in Table \ref{tab1}.
For all the 3 datasets, instruction-tuning performs best.
The two-sample t-test of the difference of means
between scores for instruction-tuning and finetuning (for the 3 datasets) have \textit{p} values $<$ 0.0001 for alpha of 0.05,
showing the results are statistically significant.
The results are competitive to the neural network approaches in \cite{mathew2021docvqa}, where they introduced \acrshort{docvqa} and included additional visual object features and fixed vocabulary to improve the results.
Compared with \acrshort{sota} BLIVA (6.24\%) \cite{hu2023bliva} zero-shot we achieve far better result on \acrshort{docvqa} as expected with finetuning (20.079\%) and even better result with instruction-tuning (20.667\%).
Compared to LoRRA (26.56\%) \cite{8953586}, where the \acrshort{tvqa} dataset was introduced, we achieve better performance (39.19\%), though not as good as BLIVA (42.18\%) zero-shot but beating most contenders like MiniGPT4, OpenFlamingo, InstructBLIP, and mPLUG-Owl with 18.72\%, 29.08\%, 36.86\%, and 37.44\% respectively.

\begin{table}[h!]
\small
\centering
\caption{Average accuracy (\%) scores (pixel only, i.e. without \acrshort{ocr}) and standard deviation. Models based on \acrshort{llava}1.5-7B. Instruction-tuning has the best performance.}\label{tab1}
\begin{tabular}{p{0.12\linewidth} | p{0.15\linewidth} | p{0.15\linewidth} | p{0.18\linewidth} | p{0.18\linewidth} | p{0.18\linewidth} }
\hline
 & \multicolumn{5}{|c}{\textbf{Model} Accuracies (\%) $\uparrow$} \\
\hline
\textbf{Data} & \textbf{Literature (LoRRA)} & \textbf{Baseline} & \textbf{Finetuning}  & \textbf{50\% tuning} & \textbf{Instruction-tuning} \\
\hline
\acrshort{docvqa} & 7.09 \cite{mathew2021docvqa} & 1.673 (0.18) & 20.079 (0.07) & 19.527 (0.18) & \textbf{20.667} (0.18) \\
 & &  &  & & \\
\hline
TextVQA  & 26.56 \cite{8953586} & 2.79 (0.13) & 38.67 (0.13) & 38.65 (0.13) & \textbf{39.19} (0.13) \\
&  &  &  & & \\
\hline
\acrshort{idvqa} & - &  0.952 (0.01) & 29.52 (0.03) & 29.438 (0.01) & \textbf{29.583} (0.01) \\
&  &  &  & & \\
\hline
\end{tabular}
\end{table}

\begin{table}[h]
\centering
\caption{Hallucination evaluation using \acrshort{pope} on \acrshort{idvqa}-based models.}\label{tab2}
\begin{tabular}{|c|c|c|c|}
\hline
 & \multicolumn{3}{|c|}{\textbf{POPE}}  \\
\hline
 & \multicolumn{3}{|c|}{\textbf{F1} $\uparrow$ | \textbf{Yes} ratio $\downarrow$ scores} \\
& Adversarial & Random & Popular \\
\hline
Baseline & 0.853 (0) | 0.469 (0) & 0.885 (0) | 0.448 (0) & 0.873 (0) | 0.448 (0) \\
 &  &  & \\
\hline
Finetuning & 0.836 (0) | 0.518 (0) & 0.88 (0) | 0.481 (0) & 0.882 (0) | 0.464 (0) \\
 &  &  &  \\
\hline
50\% tuning  & 0.833 (0) |  0.611 (0) & 0.897 (0) | 0.548 (0) & 0.884 (0) | 0.548 (0) \\
 &  &  & \\
\hline
Instruction-tuning  & 0.819 (0) | 0.5 (0) & 0.868 (0) | 0.457 (0) & 0.861 (0) | 0.452 (0) \\
 &  &  & \\
\hline
\end{tabular}
\end{table}

\begin{figure}[h]
\centering
\includegraphics[width=1\textwidth]{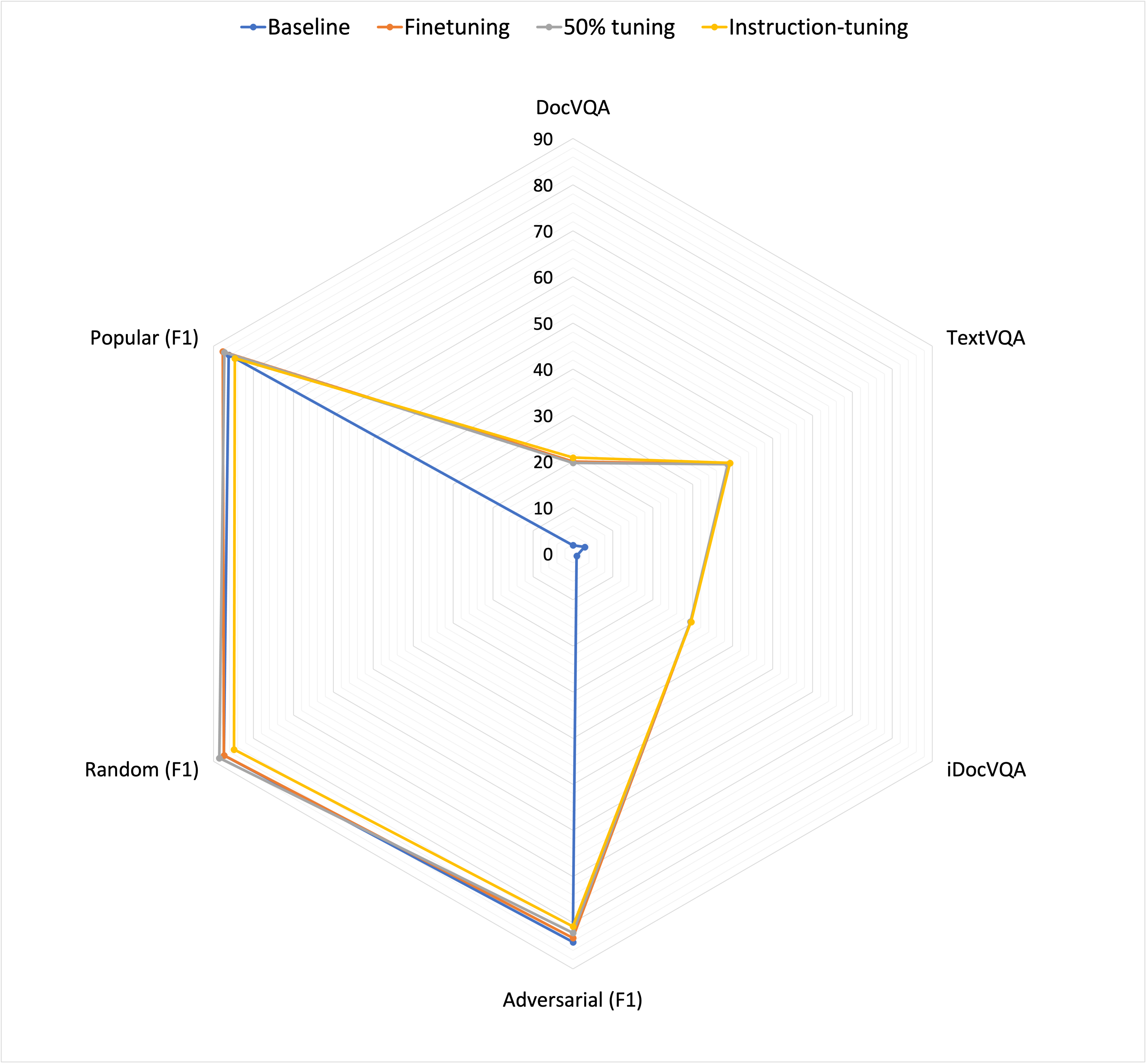}
\caption{Spider chart of the performance of the models.} \label{fig1}
\end{figure}

We also evaluated hallucination.
We observe from Table \ref{tab2} that the base model and derived models are not too over-confident because they give \textit{Yes} ratio scores below 0.62 \cite{li-etal-2023-evaluating}, leading to better F1 scores, compared to most results obtained by \cite{li-etal-2023-evaluating}, where it is introduced.
This suggests the models are less prone to hallucinations.
However, similarly to \cite{li-etal-2023-evaluating}, we also observe that (\textit{F1}) performance per model generally falls from random settings, to popular and adversarial.
Figure \ref{fig1} is a spider chart of the results.
Overall, interpreting the scores calls for standard precaution that suggests balancing accuracy performance and possible level of hallucination.
We also experimented with the parameter-efficient LoRA finetuning, but the accuracy results were slightly worse than what is provided in Table \ref{tab1}.
This also applies to smaller epoch number of 6.
Merging the two datasets to form \acrshort{idvqa} produces a broader \acrshort{vqa} challenge.
The work in this paper also provides a baseline for the performance on this new merged dataset.

\subsection*{Qualitative Results}
Figures \ref{figd2} to \ref{figidoc2} provide six examples of where the instruction-tuning models outperform other models.
Figure \ref{figt2} is an example that may be considered very challenging, even for humans.
Despite the correct examples of the instruction-tuning that are provided, there are examples where it obviously was incorrect, given its accuracy.
For instance, Figure \ref{figidoci} is an example where the instruction-tuning model is incorrect, due to some hallucination.
The incorrect responses from the finetuning examples are likely due to hallucinations also.

\begin{figure}
\centering
\includegraphics[width=1\textwidth]{images/snbx0223_9.png}
\caption{\acrshort{docvqa} example where Instruction-tuning outperforms others.\\ Q: What is the brand name of the chips/snacks produced by ITC?"\\ \textbf{Correct}: Instruction-tuning: Bingo, Incorrect: Finetuning: Tangles} \label{figd2}
\end{figure}

\begin{figure}
\centering
\includegraphics[width=1\textwidth]{images/mybv0228_13.png}
\caption{\acrshort{docvqa} example where Instruction-tuning outperforms others.\\ Q: What is the variable taken along the x axis ?\\  \textbf{Correct}: weeks of consumption, Incorrect: Finetuning: weeks} \label{figd1}
\end{figure}

\begin{figure}
\centering
\includegraphics[width=0.8\textwidth]{images/bdbfae49210475d4.jpg}
\caption{\acrshort{tvqa} example where Instruction-tuning outperforms others.\\ Q: what is the largest measurement we can see on this ruler?\\ \textbf{Correct}: Instruction-tuning: 50,  Incorrect: Finetuning: 40} \label{figt1}
\end{figure}

\begin{figure}
\centering
\includegraphics[width=0.8\textwidth]{images/7b0112f7f5f096c5.jpg}
\caption{\acrshort{tvqa} example where Instruction-tuning outperforms others.\\Q: what is the year on the calender? \\ \textbf{Correct}: Instruction-tuning: 2010, Incorrect: Finetuning: 2014} \label{figt2}
\end{figure}

\begin{figure}
\centering
\includegraphics[width=1\textwidth]{images/snbx0223_44.png}
\caption{\acrshort{idvqa} example where Instruction-tuning outperforms others.\\ Q: How many nomination committee meetings has S. Banerjee attended?\\ \textbf{Correct}: Instruction-tuning: 2,  Incorrect: Finetuning: 34} \label{figidoc1}
\end{figure}

\begin{figure}
\centering
\includegraphics[width=1\textwidth]{images/xjpn0081_4.png}
\caption{\acrshort{idvqa} example where Instruction-tuning outperforms others.\\ Q: How many grass/straw pieces of matter is found in the core samples?\\ \textbf{Correct}: Instruction-tuning: 2,  Incorrect: Finetuning: 23} \label{figidoc2}
\end{figure}

\begin{figure}
\centering
\includegraphics[width=1\textwidth]{images/nkbl0226_1.png}
\caption{\acrshort{idvqa} example where Instruction-tuning is incorrect.\\ Q: What is name of university?\\ \textbf{Correct}: university of california,  Incorrect: Instruction-tuning: university of massachusetts} \label{figidoci}
\end{figure}

\section{Conclusion}
\label{conclude}
This work has shown that instruction datasets for instruction-tuning are effective for improving the performance of \acrshort{lmm}s.
The \acrshort{docvqa} task can benefit from \acrshort{lv} models, which provide the basis for additional novel tasks besides it.
Well-crafted instructions enable the underlying \acrshort{llm} take on a helpful persona in solving difficult problems, even in the domain of \acrshort{docvqa}.
In spite of the improvements witnessed with instruction-tuning in this work, there's still a wide gap in performance when compared to humans \cite{mathew2021docvqa}.
Future work may involve evaluating the performance gains possible for the \acrshort{docvqa} task across additional \acrshort{lmm}s and to create multilingual instruction datasets for instruction-tuning.
One may also enable \acrshort{llad} to process multiple images, possibly by expanding the context size of the base model.
These steps, in addition to others, may provide improvements towards human-like performance and better generalizability.

\section{Limitations}
While we made careful effort to evaluate the challenge of hallucination of the models, it is likely that the models may still be susceptible to hallucinations.
The underlying \acrshort{llm}, \acrshort{llama}-2, may also be susceptible to generating other harmful content, given that this is a well-known challenge with \acrshort{llm}s \cite{info13060298}.


%
%
%
\bibliographystyle{splncs04}
\bibliography{mybibliography}

\end{document}